\documentclass[letterpaper]{article} 
\usepackage{aaai2026}  
\usepackage{times}  
\usepackage{helvet}  
\usepackage{courier}  
\usepackage[hyphens]{url}  
\usepackage{graphicx} 
\usepackage{natbib}  
\usepackage{caption} 
\usepackage{algorithm}
\usepackage{algorithmic}
\usepackage{amsmath}
\usepackage{multirow}

\usepackage{amssymb}   
\usepackage{booktabs}
\usepackage{xcolor}
\usepackage{threeparttable}
\usepackage{adjustbox}
\usepackage{array}
\usepackage{siunitx}
\usepackage{makecell} 
\usepackage{bm}

\usepackage{url}
\usepackage{newfloat}
\usepackage{listings}
\DeclareCaptionStyle{ruled}{labelfont=normalfont,labelsep=colon,strut=off} 
\floatstyle{ruled}
\newfloat{listing}{tb}{lst}{}
\floatname{listing}{Listing}
\title{HiMo-CLIP: Modeling Semantic Hierarchy and Monotonicity \\ in Vision-Language Alignment}
\author{
    Ruijia Wu\textsuperscript{\rm 1,\rm 2}\textsuperscript{\dag},
    Ping Chen\textsuperscript{\rm 1,\rm 2}\textsuperscript{\dag},
    Fei Shen\textsuperscript{\rm 3},
    Shaoan Zhao\textsuperscript{\rm 1,\rm 2},
    Qiang Hui\textsuperscript{\rm 1,\rm 2},
    Huanlin Gao\textsuperscript{\rm 1,\rm 2},\\
    Ting Lu\textsuperscript{\rm 1,\rm 2},
    Zhaoxiang Liu\textsuperscript{\rm 1,\rm 2},
    Fang Zhao\textsuperscript{\rm 1,\rm 2}\textsuperscript{*},
    Kai Wang\textsuperscript{\rm 1,\rm 2},
    Shiguo Lian\textsuperscript{\rm 1,\rm 2}\textsuperscript{*}
}

\affiliations{
  
    {\tt\large \textsuperscript{$^\dag$Equal contribution, $^*$Corresponding Authors }} \\
    \textsuperscript{\rm 1}Data Science \& Artificial Intelligence Research Institute, China Unicom\\
    \textsuperscript{\rm 2}Unicom Data Intelligence, China Unicom\\
    \textsuperscript{\rm 3}National University of Singapore\\
    
    {\tt\small \{wurj25, chenp181, zhaof50, liansg\}@chinaunicom.cn, shenfei29@nus.edu.sg}
}

\usepackage{bibentry}

\begin{document}

\maketitle
\begin{abstract}
Contrastive vision-language models like CLIP have achieved impressive results in image-text retrieval by aligning image and text representations in a shared embedding space. However, these models often treat text as flat sequences, limiting their ability to handle complex, compositional, and long-form descriptions. In particular, they fail to capture two essential properties of language: \textbf{semantic hierarchy}, \textit{which reflects the multi-level compositional structure of text}, and \textbf{semantic monotonicity}, \textit{where richer descriptions should result in stronger alignment with visual content}.
To address these limitations, we propose HiMo-CLIP, a representation-level framework that enhances CLIP-style models without modifying the encoder architecture. HiMo-CLIP introduces two key components: a hierarchical decomposition (HiDe) module that extracts latent semantic components from long-form text via in-batch PCA, enabling flexible, batch-aware alignment across different semantic granularities, and a monotonicity-aware contrastive loss (MoLo) that jointly aligns global and component-level representations, encouraging the model to internalize semantic ordering and alignment strength as a function of textual completeness.
These components work in concert to produce structured, cognitively-aligned cross-modal representations. Experiments on multiple image-text retrieval benchmarks show that HiMo-CLIP consistently outperforms strong baselines, particularly under long or compositional descriptions. The code is available at \textcolor{blue}{ \url{https://github.com/UnicomAI/HiMo-CLIP}}.
\end{abstract}


\begin{figure}[t]
    \centering
    \includegraphics[width=1\linewidth]{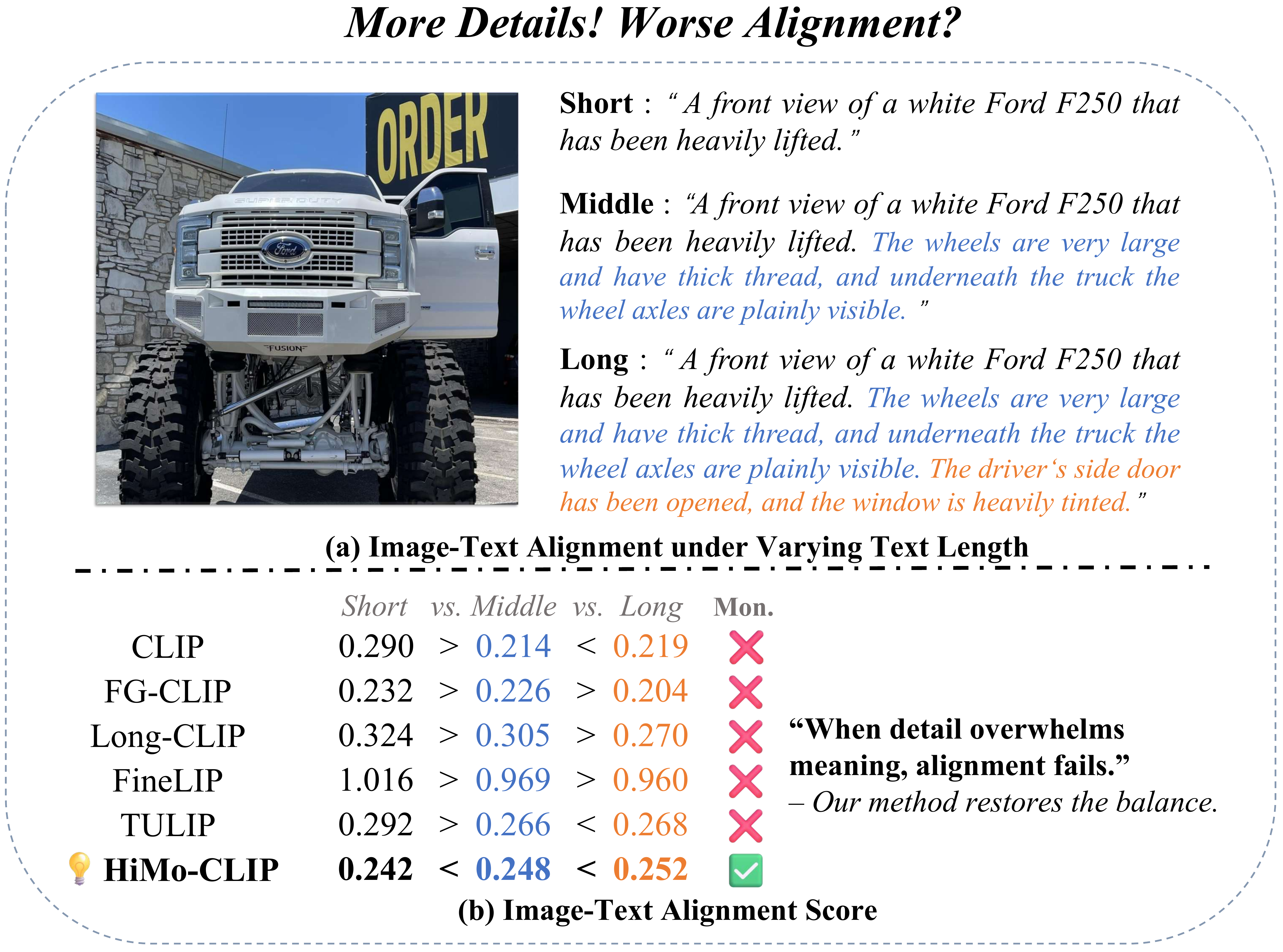}
    \caption{%
    (a) Text descriptions of an image often grow in semantic richness, from short to long, by adding more visual details. (b) However, existing models, even those tailored for long-form text, often fail to preserve semantic monotonicity, overlooking this essential principle when scaling to richer descriptions. In contrast, HiMo-CLIP maintains alignment consistency across text granularities, effectively addressing this overlooked yet critical challenge. (\textit{Note: FineLIP’s similarity exceeds 1 due to its customized test-time scaling.})%
    }
    \label{fig:intro}
    \vspace{-0.5cm}
\end{figure}

\section{1. Introduction}
\label{sec:intro}

Cross-modal contrastive learning has become a foundational approach for vision-language tasks, demonstrating strong performance in image-text retrieval, zero-shot classification, and image captioning. Models such as CLIP~\cite{radford2021learning} achieve this by projecting images and texts into a shared embedding space, where semantically aligned pairs are drawn closer while mismatched pairs are pushed apart. However, these models typically treat textual inputs as flat, unstructured sequences, ignoring the rich semantic hierarchy and compositionality inherent in natural language~\cite{lewis2022does,ji2021step,koukounas2024jina}. This simplification limits their ability to fully exploit longer or more complex descriptions that contain multi-level semantics beyond short captions~\cite{guo2023hgan,wu2021hanet}.

In practice, a single image may be described at multiple semantic levels. As shown in Figure~\ref{fig:intro}, the same image of a white Ford F250 can be paired with either a short caption, “A front view of a white Ford F250 that has been heavily lifted”, or a longer, more detailed description that elaborates on visual attributes such as the oversized wheels, visible axles, and tinted windows. A cognitively aligned model should not only assess whether a text matches the image, but also reason about how it matches, through object category, appearance, or fine-grained contextual cues~\cite{zeng2021multi,tian2025llm,sun2024alpha}. This motivates two underexplored properties in vision-language contrastive learning: \textbf{semantic hierarchy}, \textit{the ability to represent and align descriptions across varying levels of semantic granularity}, and \textbf{semantic monotonicity}, \textit{the expectation that more informative and complete descriptions should lead to stronger alignment with the corresponding image}.

Modeling these properties poses significant challenges. Existing methods often rely on fixed-length captions or handcrafted subphrases to approximate semantic granularity, yet such designs overlook the dynamic and context-dependent nature of semantic focus. As illustrated in Figure~\ref{fig:intro}, the long-form description of the Ford F250 includes multiple visual cues, such as “oversized wheels,” “visible axles,” and “tinted windows.” However, which part of the description should be most semantically aligned depends on the batch context. For instance, in a batch containing other trucks, “tinted windows” may offer the most discriminative cue, whereas in a batch with various vehicle types, “Ford F250” or “heavily lifted” may dominate the alignment. Static, substring-based decomposition, such as fixed truncation or manual segmentation, cannot flexibly adapt to such shifts, and may introduce semantic noise or supervision bias~\cite{yuksekgonul2022and}. This calls for a data-driven, context-aware mechanism to decompose and align semantic components at varying levels of granularity.

To address the above challenges, we propose HiMo-CLIP, a representation-level framework that enhances CLIP-style models by explicitly modeling semantic hierarchy and monotonicity, without modifying the underlying encoder architecture. HiMo-CLIP introduces two key components: a Hierarchical Decomposition (HiDe) module and a Monotonicity-aware contrastive Loss (MoLo). The HiDe module leverages in-batch Principal Component Analysis (PCA)~\cite{jolliffe2011principal} to extract latent semantic components from long-form text, enabling flexible, batch-aware alignment across varying semantic granularities. This allows the same sentence to emphasize different aspects, such as object type, attributes, or context-depending on the distribution of the current batch. Unlike manual subphrases or truncated captions, these components are derived in a self-supervised and context-adaptive manner, ensuring both semantic consistency and scalability during training.
The MoLo loss complements HiDe by introducing a dual-branch alignment objective: one branch aligns global image-text pairs, while the other aligns image features with each semantic component independently. This formulation encourages the model to internalize semantic ordering, i.e., more detailed and informative texts should yield stronger alignment signals. Notably, all components operate purely in the embedding space, requiring no changes to the pretrained encoder or additional supervision. Extensive experiments across multiple image-text retrieval benchmarks demonstrate that HiMo-CLIP consistently outperforms strong baselines, particularly under long-form or compositional descriptions.

Our contributions are summarized as follows:
\begin{itemize}
\item We identify and formally define \textbf{semantic hierarchy} and \textbf{semantic monotonicity} as two fundamental yet underexplored properties of cross-modal contrastive learning, essential for modeling multi-granular and completeness-sensitive text-image alignment.

\item We propose \textbf{HiMo-CLIP}, a self-supervised and encoder-agnostic framework that contains a Hierarchical Decomposition module (HiDe) and a Monotonicity-aware Contrastive loss (MoLo). Together, they enable structured, context-aware, and monotonic alignment without requiring architectural modifications or external annotations.

\item We validate HiMo-CLIP on multiple image-text retrieval benchmarks and demonstrate consistent improvements over strong CLIP-style baselines, particularly in scenarios involving long-form or compositional descriptions.
\end{itemize}

\begin{figure*}[t]
    \centering\includegraphics[width=0.9\linewidth]{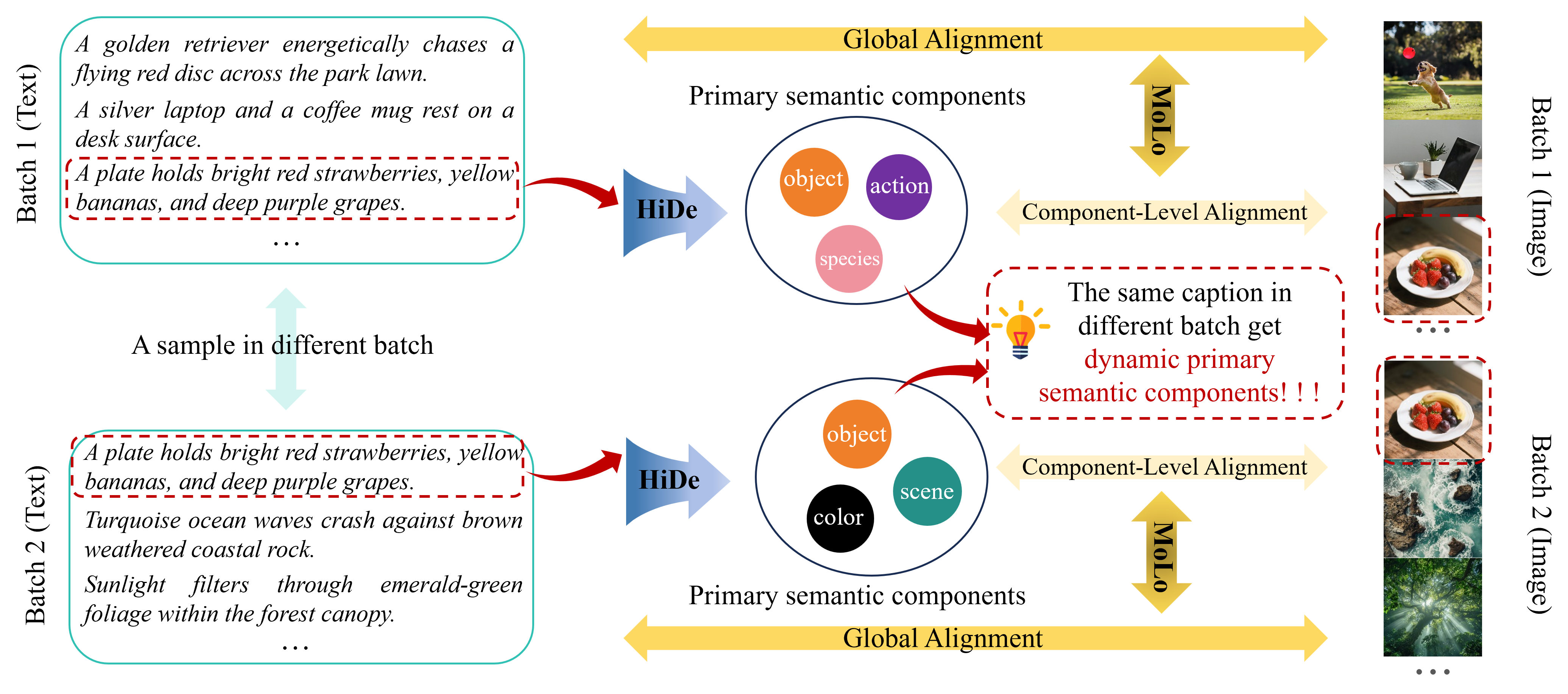}
    \caption{\textbf{HiMo-CLIP Framework.} Our method enhances CLIP with two encoder-agnostic modules: (1) \textbf{HiDe} performs in-batch PCA to extract the most discriminative semantic components from each text, adapting to batch context and revealing dynamic semantic hierarchy. For instance, the same text (red dashed box) yields different components in different batches. (2) \textbf{MoLo} enforces a dual alignment objective, aligning images with both full-text embeddings (\textit{Global Alignment}) and their primary semantic components (\textit{Component-Level Alignment}), promoting semantic monotonicity.}

    \label{fig:framework}
    \vspace{-0.5cm}
\end{figure*}

\section{2. Related Work}

\subsection{2.1 Vision-Language Pretraining with CLIP}
Vision-language pretraining has advanced rapidly with models like CLIP~\cite{radford2021learning}, which leverage contrastive learning over large-scale image-text pairs to learn transferable cross-modal representations. CLIP adopts a dual-encoder architecture, using a visual backbone (e.g., ViT) and a text encoder (e.g., Transformer) to project images and texts into a shared embedding space, where an InfoNCE loss~\cite{oord2018representation} encourages alignment of positive pairs.
Despite its strong performance in zero-shot tasks and retrieval, CLIP is limited by its reliance on short textual inputs~\cite{alper2024emergent,xu2023demystifying}. The 77-token constraint, with most information packed into the first 20 tokens, hampers its ability to encode rich, structured semantics~\cite{zhang2024long}. Long-form descriptions are often truncated, and key syntactic or compositional relationships (e.g., attributes, spatial cues) are lost or entangled with verbose modifiers~\cite{tschannen2025siglip,li2022blip}. This leads to flattened, “bag-of-words”-like embeddings that fail to reflect fine-grained semantics, restricting the model’s ability to support deeper alignment and reasoning~\cite{yamada2022lemons,eva02,zhai2023sigmoid}.
Several datasets have been designed to evaluate hierarchical structure, but they are primarily focused on short-text alignment and do not fully address the challenges of long-form text~\cite{vulic2017hyperlex, santurkar2020breeds, alper2024emergent}.

\subsection{2.2 Long-Form Text Modeling}
To extend contrastive models beyond short captions, recent works have introduced architectural modifications and alignment strategies for handling long-form text. LongCLIP~\cite{zhang2024long} increases token capacity via positional interpolation and aligns global image features with long-text backbones through principal component matching. FineLIP~\cite{asokan2025finelip} refines token-level alignment via adaptive modulation, while DreamLIP~\cite{zheng2024dreamlip} generates subcaptions to facilitate hierarchical alignment between local regions and text fragments. FG-CLIP~\cite{xie2025fg} introduces region-aware hard negatives to boost fine-grained discrimination.
Despite these advances, most methods focus on simplifying the visual side, through downsampling, cropping, or sparse selection, to accommodate longer texts, often degrading visual fidelity. More critically, few address the semantic redundancy and structural diffusion intrinsic to long-form language. This results in unstable performance when input text becomes more detailed. LoTLIP~\cite{wu2024lotlip} introduces clause-aware corner tokens to better capture long-text semantics, while TULIP~\cite{najdenkoska2024tulip} extends token limits via relative position encoding. However, both lack mechanisms for structured semantic compression, leaving redundancy and irrelevant details insufficiently filtered.

Overall, existing models tend to overlook the modality asymmetry: images typically exhibit spatial density and local coherence, whereas long-form text is verbose and hierarchically structured, highlighting the need for structured semantic compression on the language side.

\section{3. Method}

We propose HiMo-CLIP, a representation-level framework that enhances CLIP-style models to capture semantic hierarchy and monotonicity without modifying encoder architectures (see Fig.~\ref{fig:framework}). The framework introduces two core components: a Hierarchical Decomposition (HiDe) module that extracts multi-granular semantic components via in-batch PCA, and a Monotonicity-aware contrastive Loss (MoLo) that leverages these components to enforce natural semantic ordering. Operating entirely in the embedding space, HiMo-CLIP enables self-supervised learning of structured, cognitively aligned cross-modal representations.

\subsection{3.1 Overview}

HiMo-CLIP builds upon CLIP’s dual-encoder paradigm, where images and texts are projected into a shared embedding space. For an image $I$, its visual embedding is denoted by $v = f_v(I) \in \mathbb{R}^d$, and for a text $T$, the textual embedding is $u = f_t(T) \in \mathbb{R}^d$. While CLIP effectively aligns these global representations, it overlooks two essential linguistic properties: semantic hierarchy, which reflects the multi-level compositional structure inherent in natural language, and semantic monotonicity, which implies that more complete textual descriptions should result in stronger alignment with the corresponding image. HiMo-CLIP addresses these limitations by introducing two lightweight and encoder-agnostic modules. The Hierarchical Decomposition (HiDe) module dynamically extracts latent semantic components from text embeddings using in-batch Principal Component Analysis (PCA), capturing varying levels of semantic granularity. The Monotonicity-aware Contrastive Loss (MoLo) jointly aligns images with both full-text embeddings and their semantic components, implicitly encouraging alignment scores to increase as the text becomes more complete. Both modules operate entirely at the representation level, requiring no modification to the pretrained encoders, and enable structured and cognitively consistent cross-modal alignment.

\subsection{3.2 Hierarchical Decomposition (HiDe) Module}

\noindent\textbf{Motivation.} Natural language descriptions, particularly long-form ones, often express semantics across multiple levels, such as object categories, attributes, and contextual details. Capturing this inherent hierarchy is crucial for precise vision-language alignment. However, existing methods typically rely on static subphrases or fixed truncation strategies, which are inadequate for adapting to batch-dependent semantic relevance. For example, in a batch dominated by various vehicle types, category-level cues like “Ford F250” may be most salient, whereas in a batch of similar trucks, fine-grained attributes like “tinted windows” may become more informative. A decomposition strategy that is static across batches fails to reflect such context shifts, resulting in suboptimal alignment. To address this, HiDe introduces a dynamic and context-aware mechanism that leverages in-batch PCA to extract latent semantic components from text embeddings. This allows the model to capture semantically meaningful structures at varying levels of granularity, adaptively shaped by the composition of the current batch.

\noindent\textbf{Architecture.} Given a mini-batch $\mathcal{B} = \{(I_i, T_i)\}_{i=1}^N$ of $N$ image-text pairs, HiDe decomposes textual embeddings into context-aware semantic components via PCA. Each text input $T_i$ is encoded as $u_i = f_t(T_i) \in \mathbb{R}^d$. The batch mean embedding is computed as $\bar{u} = \frac{1}{N} \sum_{j=1}^N u_j$, and each embedding is centered by $\hat{u}_i = u_i - \bar{u}$. PCA is then performed on the set $\{\hat{u}_i\}_{i=1}^N$ through Singular Value Decomposition (SVD) to extract the top $m$ principal components, organized as $\mathbf{P} = [p_1, \ldots, p_m]^\top \in \mathbb{R}^{m \times d}$, where $m$ is chosen such that the cumulative explained variance exceeds a predefined threshold (e.g., 0.9). Each centered embedding is projected onto these principal components and reconstructed by
\begin{equation}
u_i' = \mathbf{P}^\top (\mathbf{P} \hat{u}_i) + \bar{u}.
\end{equation}
This procedure produces a semantic component vector $u_i'$ for each original embedding $u_i$, serving as a compact sub-semantic representation that underpins the hierarchical alignment within the HiMo-CLIP framework.

\subsection{3.3 Monotonicity-Aware Contrastive Loss (MoLo)}

\noindent\textbf{Motivation.} Semantic monotonicity captures the intuition that more complete textual descriptions should align more strongly with their images than partial ones. Standard contrastive models treat each input as an independent unit, lacking this ordering property. To address this, we propose a dual-level objective jointly optimizing global text embeddings and their semantic components. Extracted via PCA from the full text, these components inherently represent partial semantics. Aligning both levels encourages the model to learn that full-text embeddings, containing all semantic substructures, yield the highest alignment scores, thereby achieving monotonicity without extra supervision.

\noindent\textbf{Architecture.} The global objective preserves CLIP’s contrastive alignment between full-image and full-text embeddings. Let $v_i$ and $u_i$ be the visual and textual embeddings for the $i$-th image-text pair. The global loss is defined as:
\begin{equation}
\scriptsize
\mathcal{L}_{\text{global}} = \frac{1}{2N} \sum_{i=1}^N \left[ \mathcal{L}_{\text{info}}(v_i, u_i) + \mathcal{L}_{\text{info}}(u_i, v_i) \right],
\end{equation}
where $\mathcal{L}_{\mathrm{info}}(a,b)$ denotes the cosine-similarity-based contrastive loss, instantiated as the widely adopted InfoNCE loss~\cite{oord2018representation, radford2021learning}. To incorporate semantic granularity, each image embedding $v_i$ is aligned with its semantic component $u_i'$ obtained via PCA. The component-level loss is formulated as:
\begin{equation}
\label{comp_loss}
\scriptsize
\mathcal{L}_{\mathrm{comp}} = \frac{1}{2N} \sum_{i=1}^N \left[ \mathcal{L}_{\mathrm{info}}(v_i, u_i') + \mathcal{L}_{\mathrm{info}}(u_i', v_i) \right].
\end{equation}
The final MoLo loss $\mathcal{L}_{\text{MoLo}}$ combines both objectives:
\begin{equation}
\scriptsize
\mathcal{L}_{\text{MoLo}} = \mathcal{L}_{\text{global}} + \lambda \cdot \mathcal{L}_{\text{comp}},
\end{equation}
where $\lambda$ balances the influence of the component-level term. This formulation implicitly enforces semantic monotonicity by leveraging the natural inclusion property of PCA components. As the components are subsets of the full-text embedding, the model learns to associate greater semantic completeness with stronger alignment, thereby producing more cognitively consistent cross-modal representations.

\begin{figure*}[t]
    \centering
    \includegraphics[width=0.8\linewidth]{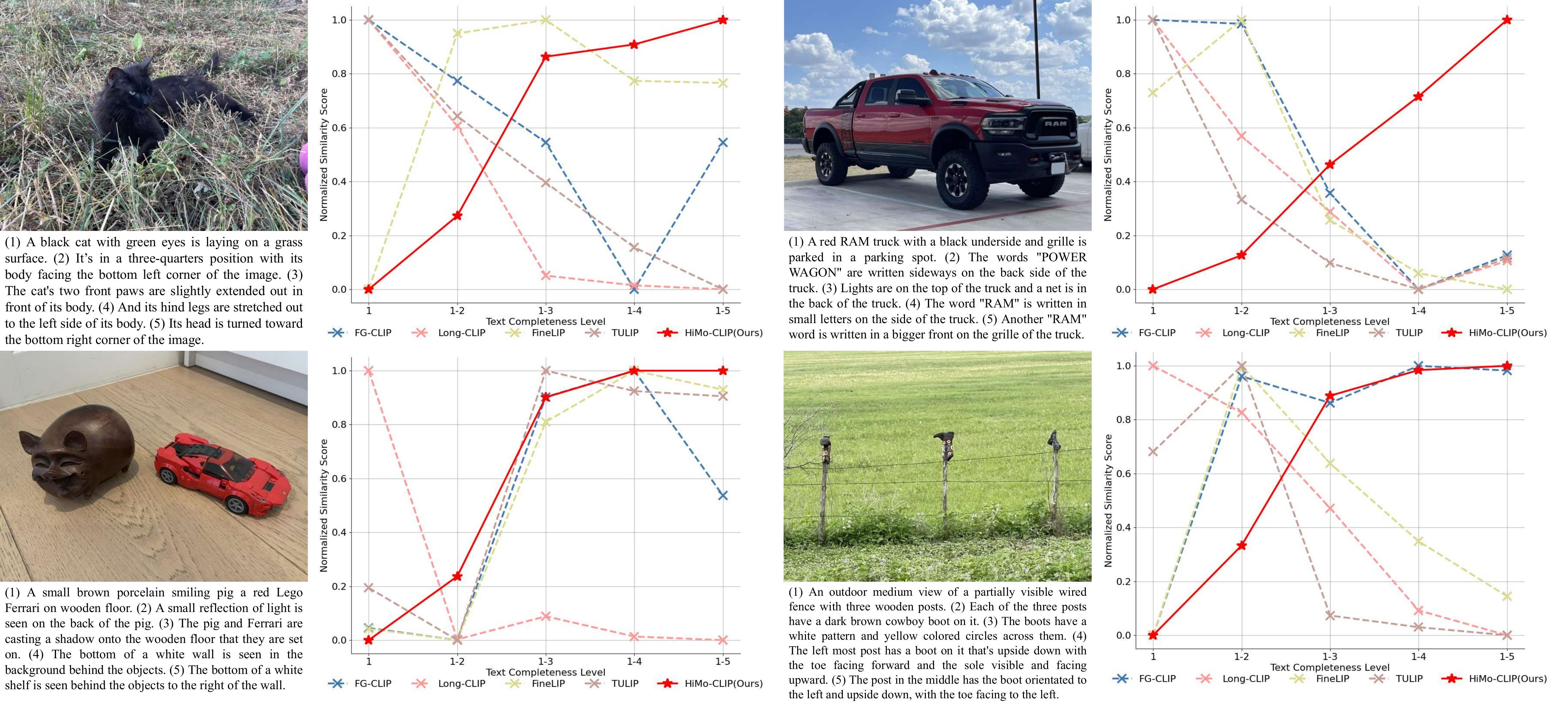}
\caption{\textbf{Semantic monotonicity on HiMo-Docci.} 
Each image is paired with five increasingly complete subtexts (HiMo@5). 
HiMo-CLIP shows consistent score increases, unlike other methods. 
All scores are sample-normalized for fair comparison.}

    \label{fig:QualityOfMonoDocci}
    \vspace{-0.5cm}
\end{figure*}

\begin{table}[t]
\centering
\caption{\textbf{Long-caption retrieval results.} 
\textasteriskcentered{} denotes reimplemented methods. 
\textbf{Bold} and \underline{underlined} indicate best and second-best scores.}
\vspace{-0.3cm}
\label{tab:long_retrieval}
\setlength{\tabcolsep}{2pt}
\renewcommand{\arraystretch}{1.05}
\begin{adjustbox}{width=\linewidth}
\begin{tabular}{l c c c c c}
\toprule
\textbf{Method} & \textbf{BackBone} & \textbf{Train Data} & \textbf{Urban1k} & \textbf{Docci} & \textbf{Long-DCI} \\
& & & I2T/T2I & I2T/T2I & I2T/T2I \\
\midrule
CLIP & ViT-B/16 & 400M & 68.1/53.6 & 58.5/58.2 & 35.0/33.1 \\
Long-CLIP & ViT-B/16 & 1M & 78.9/79.5 & 63.2/71.5 & 42.2/48.4 \\
TULIP & ViT-B/16 & 1M & 88.1/86.6 & 75.5/75.8 & 50.2/50.6 \\
FineLIP\textsuperscript{*} & ViT-B/16 & 1M & 88.2/\underline{88.2} & 75.8/77.3 & \underline{55.8}/53.1 \\
LoTLIP & ViT-B/16 & 100M & \underline{88.8}/84.8 & 73.2/71.6 & 54.5/\underline{53.3} \\
SigLIP & ViT-B/16 & 10B & 63.0/62.3 & 70.2/70.6 & 45.4/43.0 \\
BLIP & ViT-B/16 & 14M & 45.5/48.5 & 50.5/53.5 & -/- \\
EVA-02-CLIP & ViT-B/16 & 400M & 67.0/60.8 & 67.7/68.0 & -/- \\
jina-clip-v2 & ViT-B/16 & 1.7B & 80.4/78.0 & \underline{77.6}/\underline{78.2} & -/- \\
MetaCLIP & ViT-B/16 & 400M & 68.9/63.3 & 70.9/71.5 & -/- \\
\textbf{Ours} & ViT-B/16 & 1M & \textbf{89.2}/\textbf{89.6} & \textbf{77.8}/\textbf{79.9} & \textbf{58.6}/\textbf{57.1} \\
\midrule
CLIP & ViT-L/14 & 400M & 68.7/52.8 & 57.5/60.7 & 33.4/31.3 \\
Long-CLIP & ViT-L/14 & 1M & 82.7/86.1 & 66.5/78.6 & 46.5/54.3 \\
TULIP & ViT-L/14 & 1M & 90.1/91.1 & 75.5/75.8 & 55.7/56.4 \\
FineLIP\textsuperscript{*} & ViT-L/14 & 1M & \underline{92.3}/\underline{91.2} & \underline{81.5}/\underline{82.6} & \underline{59.6}/\underline{58.5} \\
MetaCLIP & ViT-L/14 & 400M & 73.4/70.0 & 76.5/76.7 & -/- \\
SAIL-L-NV2 & ViT-L/14 & 23M & 81.5/80.2 & 76.5/78.9 & -/- \\
EVA-02-CLIP & ViT-L/14 & 400M & 73.3/68.5 & 73.5/75.0 & 47.3/47.9 \\
\textbf{Ours} & ViT-L/14 & 1M & \textbf{93.0}/\textbf{93.1} & \textbf{82.4}/\textbf{84.4} & \textbf{62.2}/\textbf{61.9} \\
\bottomrule
\end{tabular}
\end{adjustbox}
\vspace{-0.5cm}
\end{table}

\section{4. Experiments}

\subsection{4.1 Training Details}
Following LongCLIP~\cite{zhang2024long}, we train HiMo-CLIP on ShareGPT4V~\cite{chen2024sharegpt4v}, which has 1.2M image–caption pairs with multi-sentence annotations averaging 143.6 words. The model is initialized from CLIP and fine-tuned 10 epochs on 8 NVIDIA H100s (global batch 1024) using AdamW ($1\mathrm{e}{-6}$ LR, 0.01 weight decay, $\beta_1=0.9$, $\beta_2=0.999$) with 200-step warm-up. Images are resized to $224\times224$, and texts padded/truncated to 248 tokens with interpolated positional embeddings. For HiDe, the explained-variance threshold is set to $\tau = 0.9$, resulting in $m$ principal components, and the component loss weight is $\lambda = 1.0$.

\subsection{4.2 Evaluation Details}

\noindent\textbf{Text-Image Retrieval.}  
We evaluate HiMo-CLIP on standard benchmarks covering image-text matching and compositional reasoning.  
For long-text retrieval, we use Urban1k~\cite{zhang2024long}, Docci~\cite{onoe2024docci}, and Long-DCI~\cite{najdenkoska2024tulip}, which contain captions averaging 128, 136, and 200 words, respectively. Evaluation follows each dataset’s official protocol using standard metrics such as Recall@K.  For compositional reasoning, we adopt COLA~\cite{ray2023cola}, which contains 210 human-validated multi-object queries with fine-grained attribute variations, and we evaluate using COLA-multi accuracy. For short-text retrieval, we report results on Flickr30k~\cite{young2014image} and COCO~\cite{lin2014microsoft}, using the same evaluation settings (please see supplementary materials).

\begin{table}[t]
\centering
\caption{\textbf{Results on short-caption cross-modal retrieval.} 
\textasteriskcentered{} indicates reimplementation.}
\vspace{-0.3cm}
\label{tab:short_retrieval}
\footnotesize
\setlength{\tabcolsep}{2pt}
\renewcommand{\arraystretch}{1.05}
\begin{adjustbox}{width=\linewidth}
\begin{tabular}{l c c c c}
\toprule
\textbf{Method} & \textbf{BackBone} & \textbf{Train Data} & \textbf{Flickr30k} & \textbf{COCO} \\
& & & I2T/T2I & I2T/T2I \\
\midrule
CLIP & ViT-B/16 & 400M & 83.3/61.9 & 51.8/32.7 \\
EVA-02-CLIP & ViT-B/16 & - & 85.7/71.2 & 58.7/42.2 \\
Long-CLIP & ViT-B/16 & 1M & 87.3/70.7 & 57.6/40.4 \\
FineLIP\textsuperscript{*} & ViT-B/16 & 1M & 85.4/65.6 & 53.1/36.2 \\
LoTLIP & ViT-B/16 & 100M & 86.9/65.2 & 59.6/38.1 \\
SigLIP & ViT-B/16 & 10B & \textbf{89.8}/\textbf{75.1} & \textbf{65.4}/\textbf{47.2} \\
Action-CLIP & ViT-B/16 & 0.7M & \underline{88.1}/\underline{74.7} & 58.4/\underline{43.2} \\
\textbf{Ours} & ViT-B/16 & 1M & \underline{88.4}/72.1 & \underline{60.6}/40.8 \\
\midrule
CLIP & ViT-L/14 & 400M & 86.1/66.0 & 56.1/35.4 \\
Long-CLIP & ViT-L/14 & 1M & 89.0/\underline{76.7} & \underline{62.8}/\underline{46.3} \\
TULIP & ViT-L/14 & 1M & 89.0/73.5 & 62.6/46.1 \\
FineLIP\textsuperscript{*} & ViT-L/14 & 1M & 90.5/74.7 & 60.3/43.5 \\
Action-CLIP & ViT-L/14-336 & 0.7M & \underline{91.5}/74.0 & 62.5/44.1 \\
\textbf{Ours} & ViT-L/14 & 1M & \textbf{92.5}/\textbf{78.2} & \textbf{65.1}/\textbf{47.2} \\
\bottomrule
\end{tabular}
\end{adjustbox}
\vspace{-0.5cm}
\end{table}

\noindent\textbf{Hierarchical Monotonic Alignment.}  
To evaluate how alignment strength scales with textual completeness, we introduce \textit{HiMo@K}, a hierarchical monotonicity metric applicable to both standard benchmarks and long-form datasets. Given a caption with $n$ sentences ($n \geq K$), we divide it into $K$ contiguous segments of roughly equal length. Let $t_k$ denote the concatenation of the first $k$ segments, and $s_{t_k}$ the matching score between the paired image and $t_k$. For each image-caption pair, we compute $\{s_{t_1}, s_{t_2}, \dots, s_{t_K}\}$ and assess whether similarity increases as more context is added. For example, with $n = 5$ and $K = 3$, the caption is split into segments as follows: the first contains sentence 1, the second sentences 2–3, and the third sentences 4–5. HiMo@3 is computed over $t_1$, $t_2$, and $t_3$ (the full caption). When $n = 6$ and $K = 2$, the two segments contain the first and second halves, respectively.

\noindent(1) \textit{General Definition ($K > 3$).}  
For deeper hierarchies, HiMo@K is defined as the Pearson correlation between subtext index $k$ and similarity score $s_{t_k}$:
\begin{equation}
\scriptsize
\text{HiMo@K} = \rho(k, s_{t_k}) = 
\frac{\sum_{k=1}^{K}(k-\bar{k})(s_{t_k}-\bar{s})}{\sqrt{\sum_{k=1}^{K}(k-\bar{k})^2} \sqrt{\sum_{k=1}^{K}(s_{t_k}-\bar{s})^2}},
\end{equation}
where $\bar{k}$ and $\bar{s}$ denote the mean of the segment indices and similarity scores.

\noindent(2) \textit{Shallow Cases ($K = 2, 3$).}  
For smaller $K$, correlation is unstable. We thus define HiMo@K as strict monotonic accuracy:
\begin{equation}
\scriptsize
\text{HiMo@K} = \frac{1}{N} \sum_{i=1}^{N} \mathbb{I}[s_{t_1}^{(i)} < s_{t_2}^{(i)} (< s_{t_3}^{(i)})] \times 100\%,
\end{equation}
where $\mathbb{I}[\cdot]$ is the indicator function, and $N$ is the number of image-caption pairs.

\noindent\textbf{Dataset for Deep Hierarchies.}  
HiMo@2 and HiMo@3 can be computed on existing datasets like Flickr30k, COCO, and Docci. However, reliable evaluation for $K > 3$ requires high-quality clause-level structure. To this end, we construct {HiMo-Docci}, a curated subset of 1,000 Docci samples with human-authored captions reannotated into semantically valid subtexts. Each caption is manually segmented and verified to ensure hierarchical granularity. Prior datasets with hierarchical structure~\cite{vulic2017hyperlex, santurkar2020breeds, alper2024emergent} focus on short text and are not well suited for long-form alignment. HiMo-Docci thus serves as a necessary resource for evaluating deep hierarchi.

\begin{table}[t]
\centering
\footnotesize 
\setlength{\tabcolsep}{1.2pt} 
\caption{\textbf{HiMo@2/3/K and COLA-multi accuracy on Urban1k, Docci, and Long-DCI.} 
\textasteriskcentered{} denotes reimplementation due to unavailable official models. 
FG-CLIP$^\dagger$ uses ViT-L/14-336; others use ViT-L/14-224. 
H2/3: HiMo@2/3; HK: HiMo@K (Pearson on HiMo-Docci); Ma.: COLA-multi accuracy. 
Ur.: Urban1k, Doc.: Docci, LD.: Long-DCI.}
\vspace{-0.3cm}
\label{tab:HiMo_metrics}
\begin{tabular}{@{}l *{4}{c} *{4}{c} c c@{}}
\toprule
\multirow{2}{*}{Method} & 
\multicolumn{4}{c}{H2$\uparrow$} & 
\multicolumn{4}{c}{H3$\uparrow$} & 
\multicolumn{1}{c}{HK$\uparrow$} & 
\multicolumn{1}{c}{Ma.$\uparrow$} \\
\cmidrule(lr){2-5} \cmidrule(lr){6-9} \cmidrule(l){10-10} \cmidrule(l){11-11}
& Ur. & Doc. & LD. & Avg & Ur. & Doc. & LD. & Avg & & \\
\midrule
CLIP          & 77.0 & 74.9 & 65.5 & 72.5 & 40.7 & 33.9 & 31.0 & 35.2 & 0.43 & 27.6 \\
EVA-02-CLIP   & 78.0 & 70.0 & 70.8 & 72.9 & 24.4 & 26.7 & 31.3 & 27.5 & 0.28 & 25.7 \\
Long-CLIP     & 34.8 & 16.7 & 53.5 & 35.0 & 39.4 & 28.5 & 41.9 & 36.6 & -0.55 & 32.4 \\
TULIP         & 97.2 & 89.2 & 83.9 & 90.1 & 62.1 & 46.6 & 45.1 & 51.3 & 0.67 & \underline{34.8} \\
FineLIP\textsuperscript{*} & \underline{97.9} & \textbf{98.5} & \underline{92.8} & \underline{96.4} & \underline{64.9} & \underline{60.6} & 53.7 & \underline{59.7} & \underline{0.83} & 34.3 \\
FG-CLIP$^\dagger$ & 96.6 & 94.1 & 91.3 & 94.0 & 62.1 & 58.5 & \underline{53.8} & 58.1 & 0.75 & 30.0 \\
\textbf{Ours} & \textbf{99.3} & \underline{98.0} & \textbf{96.4} & \textbf{97.9} & \textbf{70.9} & \textbf{62.3} & \textbf{59.3} & \textbf{64.2} & \textbf{0.88} & \textbf{38.6} \\
\bottomrule
\end{tabular}
\vspace{-0.5cm}
\end{table}

\subsection{4.3 Main Results and Analysis}
\label{sec:analysis}

\noindent\textbf{Long-form text Retrieval Performance.}
As evidenced in Table~\ref{tab:long_retrieval}, HiMo-CLIP consistently outperforms state-of-the-art methods across all long-text benchmarks. Under the \textit{ViT-L/14} backbone, our method achieves \textbf{93.0\%/93.1\%} (I2T/T2I) on Urban1k, \textbf{82.4\%/84.4\%} ((I2T/T2I)) on Docci, and \textbf{62.2\%/61.9\%} (I2T/T2I) on Long-DCI, surpassing the strongest baseline (FineLIP) by pretty margins. Notably, this is achieved while using only \textbf{1M training samples}, contrasting sharply with LoTLIP's requirement of \textbf{100M} data for competitive performance. 
TULIP and LoTLIP primarily focus on extending the text capacity without addressing semantic structure. 
TULIP's RoPE-based length extension (\textit{ViT-L/14} Docci T2I: 75.8\%) processes long texts as flat sequences, failing to compress redundant semantics. 
LoTLIP's corner tokens (\textit{ViT-B/16} Docci T2I: 71.6\%) statically bind to predefined clauses, preventing dynamic reweighting of semantic components across batches. 
In contrast, HiDe's batch-adaptive PCA (Ours: \textbf{84.4\%}) dynamically isolates context-critical elements, for example, emphasizing \textit{"tinted windows"} in vehicle-dense batches while suppressing it in scenes with varied objects.

\noindent\textbf{Short-text Retrieval Compatibility.}
Contrary to methods that sacrifice short-text performance for long-text gains (e.g., Long-CLIP's regression on Flickr30k T2I), Table~\ref{tab:short_retrieval} shows HiMo-CLIP maintains competitive results on COCO and Flickr30k. With ViT-L/14, we achieve new SOTA on Flickr30k I2T R@1 (92.5\%) and COCO T2I R@1 (47.2\%), proving our hierarchical approach \textit{generalizes} across textual granularities without overfitting to long descriptions.

\begin{table}[t]
\small
\centering
\caption{\textbf{Ablation of explained-variance threshold $\tau$ in HiDe.} 
HiMo@2: average monotonic accuracy on Urban1k, Docci, and Long-DCI. 
HiMo@K: Pearson correlation on HiMo-Docci.}
\label{tab:ablation_cumratio}
\vspace{-0.3cm}
\begin{adjustbox}{width=\linewidth}
\begin{tabular}{@{}c c c c c c@{}}
\toprule
\multirow{2}{*}{$\tau$} &
\multicolumn{1}{c}{Urban1k} &
\multicolumn{1}{c}{Docci} &
\multicolumn{1}{c}{Long-DCI} &
\multirow{2}{*}{HiMo@2$\uparrow$} &
\multirow{2}{*}{HiMo@K$\uparrow$} \\
\cmidrule(lr){2-4}
& I2T/T2I & I2T/T2I & I2T/T2I & & \\
\midrule
0.60 & 85.2/84.3 & 74.3/75.1 & 51.4/49.4 & \underline{98.2} & \underline{0.86} \\
0.70 & 91.7/92.1 & 81.3/82.8 & 60.1/59.3 & \textbf{98.3} & \underline{0.86} \\
0.80 & \underline{93.3}/\textbf{93.6} & \underline{82.3}/\underline{83.7} & \textbf{62.5}/\textbf{61.9} & 94.8 & 0.79 \\
0.85 & \textbf{93.5}/\underline{93.4} & 81.6/83.3 & 61.7/61.3 & 96.4 & 0.85 \\
0.90 & 93.0/93.1 & \textbf{82.4}/\textbf{84.4} & \underline{62.2}/\textbf{61.9} & 97.9 & \textbf{0.88} \\
0.95 & 91.7/92.8 & 80.9/83.0 & 60.8/60.4 & 94.6 & 0.81 \\
\bottomrule
\end{tabular}
\end{adjustbox}
\vspace{-0.5cm}
\end{table}

\noindent\textbf{Semantic Hierarchy and Monotonicity.}
Table~\ref{tab:HiMo_metrics} provides evidence that HiMo-CLIP addresses the limitations outlined in Section~1. Our method achieves near-perfect HiMo@2 (97.9\%) and strong HiMo@3 (64.2\%) scores, outperforming FineLIP (96.4\%, 59.7\%) and TULIP (90.1\%, 51.3\%). On deeper hierarchies (HiMo@K on HiMo‑Docci), HiMo‑CLIP achieves a Pearson correlation of 0.88, demonstrating consistent similarity growth with increasing text completeness, aligning with our semantic monotonicity objective. LongCLIP, while also using full-text supervision, trails behind (HiMo@2: 35.0\%, HiMo@3: 36.6\%) and shows a negative HiMo@K score ($-0.55$), indicating limited semantic hierarchy awareness. 
In the COLA Multi-object retrieval task, HiMo‑CLIP achieves the highest top-1 accuracy (38.6\%), outperforming TULIP (34.8\%), FineLIP (34.3\%), and LongCLIP (32.4\%). This gain highlights HiDe’s effectiveness in extracting discriminative semantics, allowing MoLo to capture fine-grained attribute-object bindings and overcome the limitations of flat sequence models, validating our hierarchical design.

\begin{figure*}[t]
    \centering\includegraphics[width=0.95\linewidth]{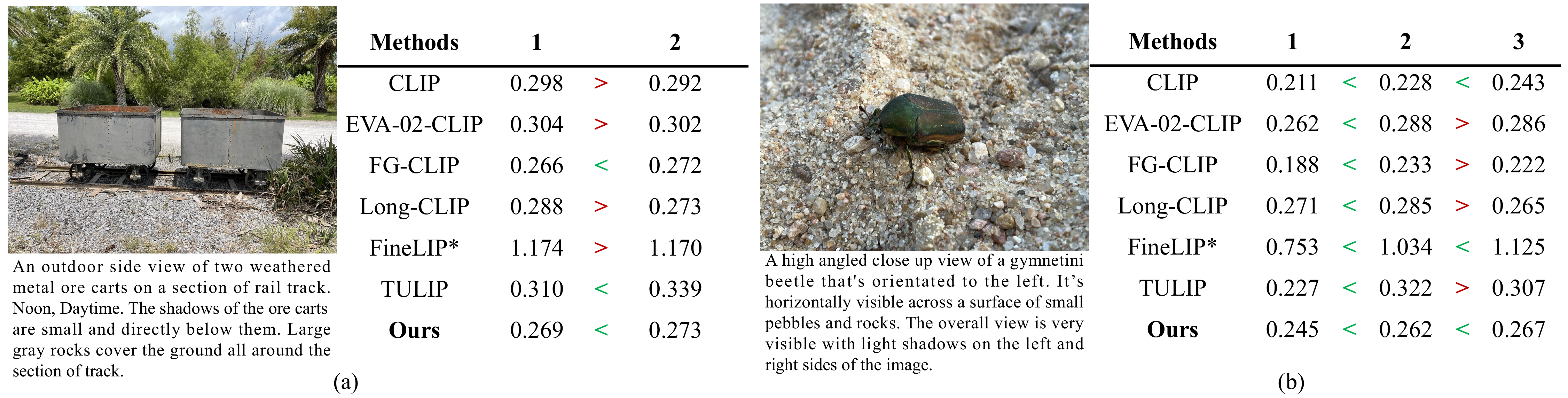}
    \caption{\textbf{Semantic monotonicity across HiMo@2 and HiMo@3 tasks.}  
(a) \textbf{HiMo@2}: Each image is paired with two text segments of increasing completeness, evaluating whether richer descriptions lead to stronger alignment.  
(b) \textbf{HiMo@3}: The image is matched with three hierarchical segments (short, medium, long), following the setup in Fig.~\ref{fig:intro}, where alignment should grow with textual detail.  
Green/red indicates correct/incorrect semantic monotonicity.  
FineLIP scores may exceed 1 due to its official score fusion strategy.
    }
    \vspace{-0.3cm}
    \label{fig:himo@23}
\end{figure*}

\begin{figure*}[t]
\centering\includegraphics[width=0.9\linewidth]{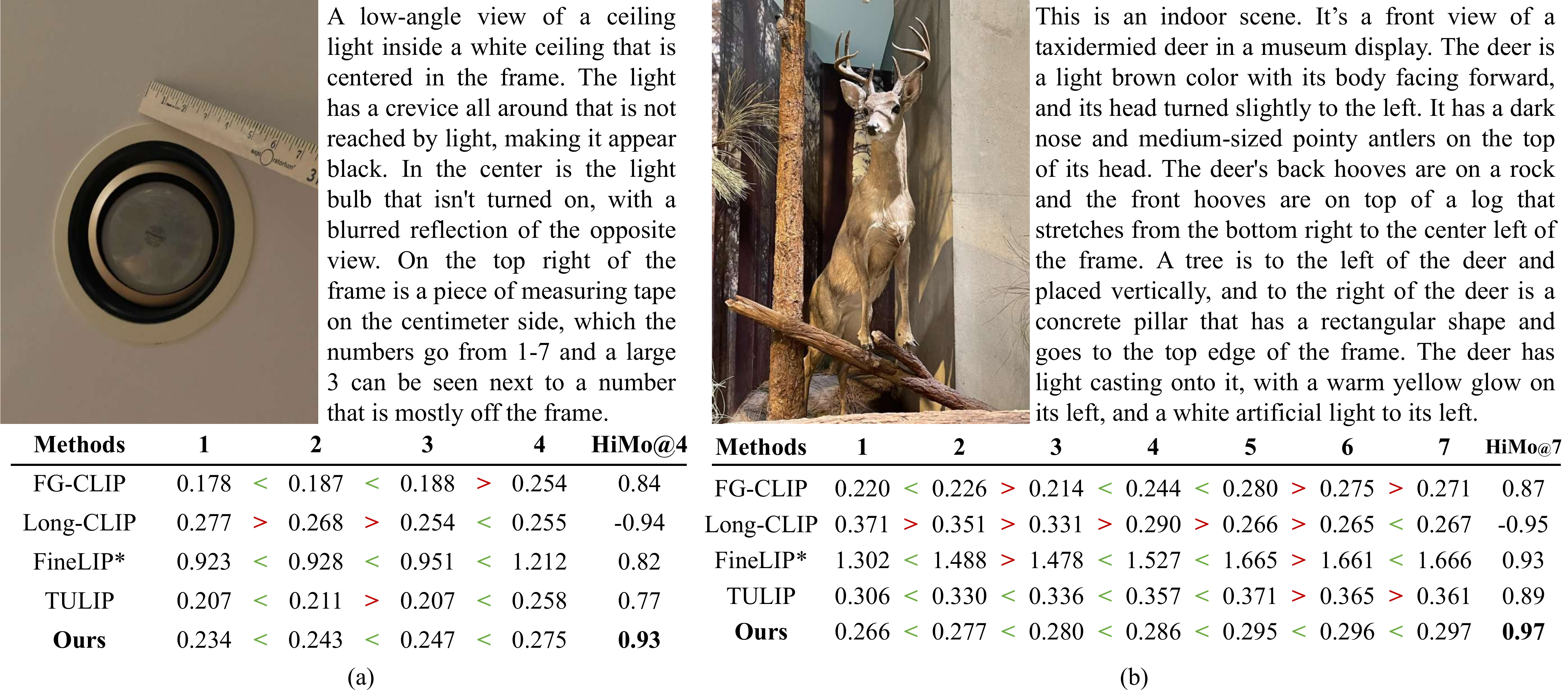}
    \caption{\textbf{Semantic monotonicity results on extended HiMo@K tasks.} 
    (a) \textbf{HiMo@4}: Each image is paired with 4 subtexts of increasing semantic detail for finer-grained monotonicity testing. 
    (b) \textbf{HiMo@7}: Each image is paired with 7 increasingly rich captions, enabling stricter evaluation of hierarchical alignment than Fig.~\ref{fig:himo@23}. 
    Green/red markers indicate correct/incorrect semantic orderings. 
    FineLIP scores may exceed 1 due to its fusion strategy.
    }
\vspace{-0.5cm}
    \label{fig:HiMo_4vs7}

\end{figure*}

\noindent\textbf{Qualitative Results.}
Figure~\ref{fig:QualityOfMonoDocci} visualizes HiMo@5 trends on HiMo-Docci, where HiMo-CLIP consistently maintains monotonic similarity growth, unlike CLIP and Long-CLIP which often exhibit erratic drops, validating our core assumption that richer subtexts should yield stronger alignment. Figure~\ref{fig:himo@23} and Figure~\ref{fig:HiMo_4vs7} extend this analysis with concrete examples for HiMo@2, @3, @4, and @7, showing that HiMo-CLIP reliably preserves correct score orderings even under deeper hierarchies. For instance, HiMo-CLIP achieves the highest qualitative HiMo@4 (0.93) and HiMo@7 (0.97), while FineLIP and TULIP exhibit score reversals, and Long-CLIP yields negative Pearson correlations ($-0.94$, $-0.95$). On shallower tasks, HiMo-CLIP maintains correct ordering at all steps, while FineLIP and TULIP show violations in HiMo@2 and HiMo@3, and even FG-CLIP fails on HiMo@3 despite strong quantitative scores. These results highlight the robustness and scalability of our representation-level alignment in modeling hierarchical semantic consistency across varied depths and content.

\subsection{4.4 Ablation Studies}

\noindent\textbf{HiDe Threshold Sensitivity.}  
As shown in Table~\ref{tab:ablation_cumratio}, setting $\tau{=}0.9$ yields the best trade-off between retrieval and hierarchy performance, with peak Urban1k/Docci recall, HiMo@2 = 97.9\%, and HiMo@K = 0.88. 
Smaller $m$ values discard essential semantic components, leading to under-representation, while larger values retain excessive noise that hampers discriminative learning. 
This confirms the importance of extracting compact and highly discriminative components through HiDe. 
These results indicate that a well-chosen decomposition threshold is crucial for capturing hierarchical semantics without weakening alignment signals essential for monotonicity.

\noindent\textbf{Component Alignment Strategy.}  
Table~\ref{tab:ablation_of_PCA} demonstrates that only our full approach ($\mathcal{L}_{\text{global}}+\mathcal{L}_{\text{comp}}$) achieves optimal results. The 0.19 gap on HiMo@K between $\mathcal{L}_{\text{global}}$ (0.69) and our method proves that joint global-component alignment is essential for monotonicity, validating MoLo's dual-branch design and confirming that naively compressed representations cannot preserve semantic ordering.  
This further validates that component-level alignment not only enriches fine-grained supervision, but also reinforces the semantic consistency across hierarchical substructures.

\noindent\textbf{Loss Weight Ablation.}  
As shown in Table~\ref{tab:ablation_lambda}, setting $\lambda{=}1$ yields the best trade-off, achieving strong global alignment (Urban1k T2I 93.1\%) and monotonicity (HiMo@2 97.9\%). Larger $\lambda$ (2) overweights components and degrades retrieval (Long-DCI I2T drops to 61.6\%), while smaller $\lambda$ (0.5) weakens monotonic structure (HiMo@2 97.1\% vs. 97.9\%).  These findings emphasize the importance of balancing global and local alignment objectives to optimize both retrieval quality and hierarchical structure.
Overall, the results demonstrate that $\lambda{=}1$ provides a balanced supervision signal, allowing the model to simultaneously retain strong retrieval performance and preserve semantic monotonicity across levels.

\begin{table}[t]
\small
\centering
\caption{\textbf{Ablation of loss and HiDe variants.} 
$\mathcal{L}_{\text{global}}$: no HiDe. 
$\mathcal{L}_{\text{comp}}$: HiDe on text only. 
$\mathcal{L}^{u,v}_{\text{comp}}$: HiDe on both modalities. 
HiMo@2: average on three long-form text datasets; HiMo@K: Pearson correlation on HiMo-Docci.}
\label{tab:ablation_of_PCA}
\vspace{-0.3cm}
\begin{adjustbox}{width=\linewidth}
\begin{tabular}{@{}l c c c c c c@{}}
\toprule
\multirow{2}{*}{Method} &
\multicolumn{1}{c}{Urban1k} &
\multicolumn{1}{c}{Docci} &
\multicolumn{1}{c}{Long-DCI} &
\multirow{2}{*}{HiMo@2$\uparrow$} &
\multirow{2}{*}{HiMo@K$\uparrow$} \\
\cmidrule(lr){2-4}
& I2T/T2I & I2T/T2I & I2T/T2I & & \\
\midrule
$\mathcal{L}_{\text{global}}$ & 91.7/92.8 & \underline{80.9}/\underline{83.0} & \underline{60.8}/60.4 & 91.0 & 0.69 \\
$\mathcal{L}_{\text{comp}}$ & \underline{92.6}/\underline{92.9} & \underline{80.9}/82.5 & 60.1/\underline{60.7} & \underline{96.2} & \underline{0.84} \\
$\mathcal{L}_{\text{global}}+\mathcal{L}^{u,v}_{\text{comp}}$ & 87.5/91.4 & 78.8/82.1 & 57.7/57.8 & 91.7 & 0.69 \\
$\mathcal{L}_{\text{global}}+\mathcal{L}_{\text{comp}}$ & \textbf{93.0}/\textbf{93.1} & \textbf{82.4}/\textbf{84.4} & \textbf{62.2}/\textbf{61.9} & \textbf{97.9} & \textbf{0.88} \\
\bottomrule
\end{tabular}
\end{adjustbox}
\vspace{-0.6cm}
\end{table}

\begin{table}[t]
\small
\centering
\caption{\textbf{Loss weight $\lambda$ ablation.} 
$\lambda$ balances global and compositional losses. 
HiMo@2: average on three datasets; HiMo@K: Pearson correlation on HiMo-Docci.}
\vspace{-0.2cm}
\label{tab:ablation_lambda}
\begin{adjustbox}{width=\linewidth}
\begin{tabular}{@{}c c c c c c@{}}
\toprule
\multirow{2}{*}{$\lambda$} &
\multicolumn{1}{c}{Urban1k} &
\multicolumn{1}{c}{Docci} &
\multicolumn{1}{c}{Long-DCI} &
\multirow{2}{*}{HiMo@2$\uparrow$} &
\multirow{2}{*}{HiMo@K$\uparrow$} \\
\cmidrule(lr){2-4}
& I2T/T2I & I2T/T2I & I2T/T2I & & \\
\midrule
2.0   & \textbf{93.1}/92.3 & 81.9/83.9 & 61.6/61.4 & 97.8 & \textbf{0.88} \\
1.0   & 93.0/\textbf{93.1} & \textbf{82.4}/\textbf{84.4} & 62.2/\textbf{61.9} & \textbf{97.9} & \textbf{0.88} \\
0.5 & 92.8/\textbf{93.1} & 82.2/84.3 & \textbf{62.3}/61.8 & 97.1 & 0.87 \\
\bottomrule
\end{tabular}
\end{adjustbox}
\vspace{-0.5cm}
\end{table}

\section{5. Conclusion}
\label{sec:conclusion}

We introduced HiMo-CLIP, a representation-level framework that enhances CLIP-style models by explicitly modeling semantic hierarchy and enforcing monotonic alignment between text and image. Our method combines a hierarchical decomposition (HiDe) module with a monotonicity-aware contrastive Loss (MoLo), enabling fine-grained alignment between image features and progressively structured subtexts. Without altering the underlying encoder architecture, HiMo-CLIP significantly outperforms prior methods on long-form text retrieval, compositional reasoning, and newly proposed HiMo@K metrics, achieving consistent gains across both qualitative and quantitative evaluations. These improvements demonstrate that aligning cross-modal representations at multiple levels of semantic abstraction is essential for robust and cognitively aligned vision-language understanding. We hope this work opens up new directions toward more structured, interpretable, and hierarchy-aware multimodal learning.

\bibliography{aaai2026}

\clearpage 
\appendix
\section*{Supplementary Material} 

\section*{Appendix A: Theoretical Perspective on HiMo-CLIP's Hierarchical Semantic Modeling}
\label{himo_theoretical}

We provide a unified theoretical analysis of the HiMo-CLIP framework, demonstrating how the \textbf{HiDe} module extracts high-level semantic components and how this decomposition naturally induces \textbf{semantic monotonicity} in image-text alignment.

\subsection*{A.1 HiDe Extracts High-Level Semantic Components}

\paragraph{Setup.}
Let a mini-batch of texts be denoted by:
\begin{equation}
\mathcal{B} = \{T_1, T_2, \ldots, T_N\}.
\end{equation}
Each text \(T_i\) is encoded into \(u_i \in \mathbb{R}^d\) via a pretrained text encoder \(f_t(\cdot)\), forming the embedding matrix:
\begin{equation}
U =
\begin{bmatrix}
u_1^\top \\
u_2^\top \\
\vdots \\
u_N^\top
\end{bmatrix}
\in \mathbb{R}^{N \times d}.
\label{eq:U_matrix}
\end{equation}
We perform mean-centering:
\begin{equation}
\hat{U} = U - \bar{u}, \quad \text{where } \bar{u} = \frac{1}{N} \sum_{i=1}^N u_i .
\label{eq:U_centered}
\end{equation}
PCA is applied to \(\hat{U}\) via SVD:
\begin{equation}
\hat{U} = U_{\text{svd}} \, \Sigma \, V^\top .
\label{eq:svd}
\end{equation}
Let \(P = V_{1:m}^\top \in \mathbb{R}^{d \times m}\) be the top-\(m\) principal directions. The projected vector is:
\begin{equation}
u_i' = P P^\top (u_i - \bar{u}) + \bar{u} .
\label{eq:project_reconstruct}
\end{equation}

\paragraph{Assumption 1: Hierarchical Semantic Composition.}
Each text embedding decomposes as:
\begin{equation}
u_i = s_i^{(1)} + s_i^{(2)} + \cdots + s_i^{(K)}, \quad s_i^{(k)} \in \mathbb{R}^d .
\label{eq:semantic_decomp}
\end{equation}
where \(s_i^{(1)}\) represents high-level semantics (e.g., category), and \(s_i^{(K)}\) low-level ones (e.g., syntax). Assume semantic orthogonality:
\begin{equation}
\mathbb{E}[\langle s_i^{(k)}, s_i^{(j)} \rangle] \approx 0, \quad \text{for } k \ne j .
\label{eq:orthogonality}
\end{equation}

\paragraph{Assumption 2: High-Level Semantics Dominate Variance.}
We define the variance of each semantic layer as:
\begin{equation}
\begin{split}
\text{Var}(k) &= \text{Tr} \left( \text{Cov}(s_1^{(k)}, \ldots, s_N^{(k)}) \right), \\
\text{where} \quad \text{Var}(1) &> \cdots > \text{Var}(K)
\end{split}
\label{eq:variance_order}
\end{equation}
Since principal component analysis (PCA) identifies directions of maximal variance, the resulting subspace \(\text{Im}(P)\) is primarily spanned by the leading semantic components \(s^{(1)}, \ldots, s^{(m^*)}\) with the highest variance.
\begin{equation}
\text{Im}(P) \subseteq \text{Span}\left(\bigcup_{k=1}^{m^*} s^{(k)}\right), \quad m^* < K .
\label{eq:projection_span}
\end{equation}
Thus, the projection \(u_i'\) approximately retains high-level semantics:
\begin{equation}
u_i' \approx s_i^{(1)} + \cdots + s_i^{(m^*)} ,
\label{eq:semantic_approx}
\end{equation}
i.e., HiDe compresses text embeddings while preserving semantically salient components and filtering noise.

\begin{table*}[t]
\centering
\small
\caption{\textbf{Comparison between Long-CLIP and HiMo-CLIP.} HiMo-CLIP achieves better semantic decomposition, monotonicity, and performance with simpler input and design.}
\begin{tabular}{p{3cm}|p{6cm}|p{6cm}}
\toprule
\textbf{Aspect} & \textbf{Long-CLIP} & \textbf{HiMo-CLIP} \\
\midrule
\textbf{Theoretical Basis} & Does not explicitly handle text redundancy; relies on short text to supplement semantics. Long text redundancy interferes with alignment. & Actively compresses textual redundancy: applies PCA to long-text features to retain core semantic components and align with linguistic hierarchies. \\
\midrule
\textbf{Sub-Semantic Extraction Logic} & PCA components of image features hardly correspond to sub-semantics (due to high-density, low-redundancy visual signals); short text often lacks key descriptions (e.g., color, attributes), leading to semantic gaps. & PCA on text naturally corresponds to sub-semantics: long text inherently contains hierarchical semantics (e.g., "Ford F250 $\rightarrow$ tinted windows"), and PCA dynamically extracts key components to avoid missing descriptions. \\
\midrule
\textbf{Support for Hierarchical Monotonicity} & No explicit mechanism; unstable alignment between short text and image subspaces (HiMo@K: -0.55). & MoLo loss implicitly enforces monotonicity: global alignment $>$ component alignment (PCA components are subsets of the full text). Theoretical proof in Appendix A\ref{himo_theoretical}(similarity increases monotonically with textual completeness). \\
\midrule
\textbf{Implementation} & \textbf{PCA Target}: Image features $\rightarrow$ InfoNCE with short text. & \textbf{PCA Target}: Text features $\rightarrow$ InfoNCE with images. \\
\midrule
\textbf{Contrastive Loss Design} & Dual branches: (i) Image vs. full long text for global alignment; (ii) PCA-image vs. short text for local alignment. & Dual branches: (i) Image vs. full long text for global alignment; (ii) Image vs. PCA-text components for local alignment. \\
\midrule
\textbf{Training Requirement} & \textbf{Text Input}: Requires both long and manually crafted short text. & Only requires raw long text; no additional annotations needed. \\
\midrule
\textbf{Computation Cost} & Higher encoding and alignment cost due to dual text types. & Single-text input; cost comparable to Long-CLIP (PCA computation is negligible). \\
\midrule
\textbf{Experimental Results} & \textbf{Semantic Monotonicity}: Significant fluctuation (HiMo@K: -0.55); alignment quality degrades with longer text. & Strict monotonic growth (HiMo@K: 0.88); similarity increases steadily with textual completeness. \\
\cmidrule{2-3}
& \textbf{Long-text Retrieval}: ViT-L/14 on Docci T2I R@1: 78.6\%. & ViT-L/14 on Docci T2I R@1: 84.4\% (+5.8\%); Urban1k/Long-DCI improves by over 7\%. \\
\bottomrule
\end{tabular}
\label{tab:himo_long_comparison}
\end{table*}

\subsection*{A.2 Monotonic Alignment via Semantic Accumulation}

Let an image-text pair be \((v, u)\), where \(v \in \mathbb{R}^d\) is the image embedding, and \(u\) is the full text embedding. Based on HiDe, we treat \(u' = u_i'\) as the compressed semantic core. Let enriched representations be defined as:
\begin{equation}
\begin{split}
u^{(1)} &= u', \\
u^{(2)} &= u' + r_1, \\
&\;\;\vdots \\
u^{(K)} &= u' + \sum_{j=1}^{K-1} r_j = u ,
\end{split}
\label{eq:residual_composition}
\end{equation}
where:
\begin{itemize}
\item \(u'\): high-level semantics from HiDe;
\item \(r_j\): semantic increment (e.g., detail, context), with \(\langle v, r_j \rangle > 0\);
\item \(|r_j|\) is non-decreasing, and directions are consistent.
\end{itemize}

Then each intermediate representation is:
\begin{equation}
u^{(k)} = u' + \sum_{j=1}^{k-1} r_j ,
\label{eq:partial_reconstruction}
\end{equation}
and the cosine similarity between image and text is:
\begin{equation}
\cos(v, u^{(k)}) = \frac{\langle v, u' \rangle + \sum_{j=1}^{k-1} \langle v, r_j \rangle}{\left\| u' + \sum_{j=1}^{k-1} r_j \right\|} .
\label{eq:cosine_reconstruction}
\end{equation}
Since the numerator increases (each increment aligns positively), and the denominator grows slowly, we obtain:
\begin{equation}
\cos(v, u^{(1)}) < \cos(v, u^{(2)}) < \cdots < \cos(v, u^{(K)}) .
\label{eq:cosine_monotonicity}
\end{equation}

\paragraph{Conclusion.}
HiDe induces a layered decomposition of semantics. When combined with MoLo, which aligns both core and enriched text representations, the framework establishes a \textbf{monotonic relationship between semantic completeness and visual alignment}:
\begin{equation}
\text{More complete text} \ \Rightarrow \ \text{Stronger alignment}
\label{eq:semantic_monotonicity}
\end{equation}

This justifies HiMo-CLIP’s ability to model \textbf{semantic monotonicity} naturally, without requiring external ranking supervision.

\begin{figure*}[t]
    \centering
    \includegraphics[width=\linewidth]{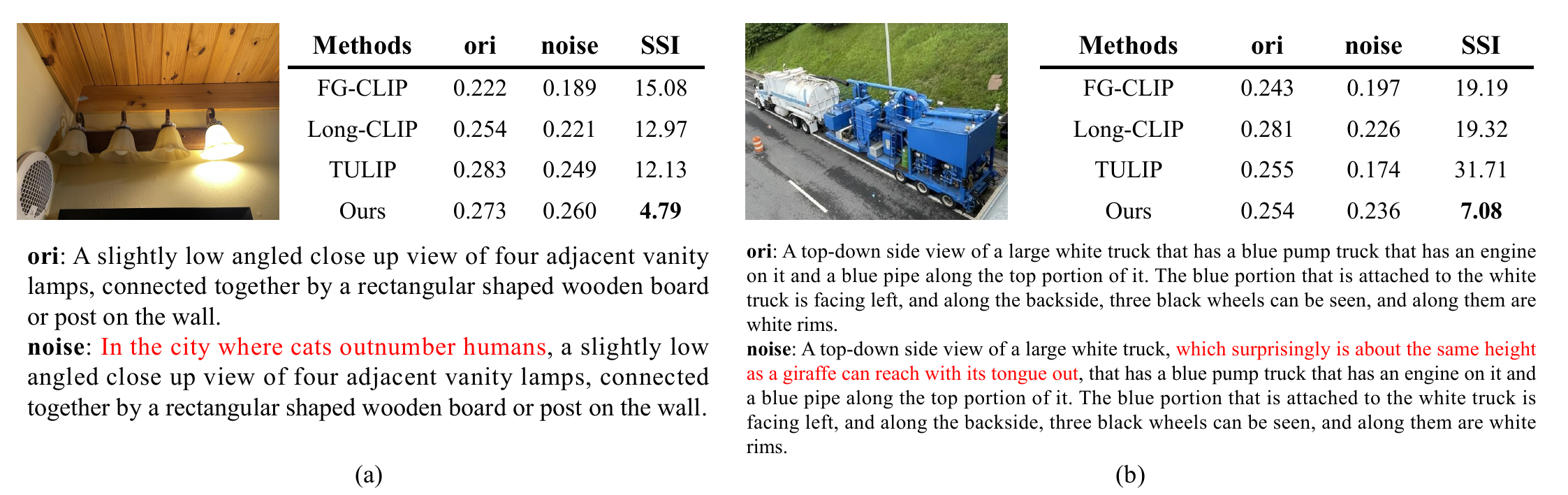}
    \caption{\textbf{Case study: robustness to semantic noise via text injection.} 
    Each example shows an image paired with a caption, into which an unrelated sentence (highlighted in red) is inserted at either the beginning or end to simulate semantic noise. 
    \textbf{Case a} describes a vanity light fixture in an indoor setting; \textbf{Case b} depicts a large blue industrial vehicle. 
    For each method, we report the image-text similarity scores before and after noise injection, as well as the corresponding Semantic Stability Index (SSI) as defined in Section~\ref{sec:robustness}. 
    HiMo-CLIP demonstrates the smallest score drop and lowest SSI in both cases (4.79 and 7.08), indicating high resilience to off-topic text. In contrast, TULIP and Long-CLIP suffer significant degradation (e.g., TULIP’s SSI reaches 31.71 in Case b), showing greater vulnerability to irrelevant content. These qualitative results support HiMo-CLIP’s ability to suppress semantically unaligned distractors through batch-aware decomposition and monotonic alignment objectives.}    
    \label{fig:ssi_cases}
\end{figure*}

\section*{Appendix B: Discussion on HiMo-CLIP and LongCLIP}
Long-CLIP and HiMo-CLIP (Our) both enhance CLIP for long-text alignment but differ fundamentally in modality processing and semantic modeling (see Table~\ref{tab:himo_long_comparison}).
\textbf{Modality Handling.} Long-CLIP decomposes image features to match short texts, overlooking the compactness of visual signals and the redundancy in language. It requires handcrafted short texts, often missing key attributes. In contrast, HiMo-CLIP compresses long-text features via PCA, adaptively extracting sub-semantics without supervision.
\textbf{Monotonicity.} Long-CLIP lacks explicit constraints and shows degraded performance as text grows (HiMo@K: -0.55). HiMo-CLIP enforces monotonic similarity via MoLo loss, leveraging PCA inclusion (HiMo@K: 0.88).

\textbf{Efficiency.} HiMo-CLIP eliminates the need for short text construction and matches Long-CLIP in cost. It achieves stronger results in retrieval (+5.8\% on Docci) and reasoning (+4.3\% on COLA), while remaining compatible with short-text tasks.

\textbf{Summary.} HiMo-CLIP revisits modality asymmetry by compressing redundant text instead of images, ensuring both efficiency and performance gains.

\section*{Appendix C: Additional Experimental Details and In-depth Analysis}
\label{himo_Experimental}

\subsection*{C.1 Inference without HiDe}
\label{sec:inference details}

During inference, we follow the standard CLIP pipeline without using HiDe or MoLo. We independently encode images and texts with the pretrained encoders, normalize their features, and compute cosine similarity for both image-to-text and text-to-image retrieval.

In contrast, FineLIP employs a more complex inference involving weighted combinations of coarse- and fine-grained similarities. Despite this simplicity, our method achieves better retrieval accuracy with lower computational overhead.

\subsection*{C.2 Robustness of HiDe to Batch Diversity}
\label{sec:batch size}

\noindent\textbf{Motivation.}  
HiDe relies on the semantic diversity within each batch to extract representative high-level components from long-form text. Larger batches naturally offer richer contextual variety, enabling HiDe to identify more stable and informative semantic structures. We thus investigate how this diversity, controlled via batch size, influences downstream performance. All experiments were conducted under our default maximum batch size of 1024 due to hardware constraints.

\noindent\textbf{Results and Analysis.}  
Table~\ref{tab:batch_size_comparison} reports performance under batch sizes of 256, 512, and 1024. Across retrieval and monotonicity metrics, we observe consistent improvements as batch size increases. For example, HiMo@2 improves from 97.3 to 97.9, and Docci-T2I from 81.8 to 84.4. This trend confirms that HiDe benefits from increased semantic variety, enabling stronger alignment with visual representations.
However, the performance gains diminish as batch size grows, indicating that HiDe remains robust even when semantic diversity is moderate. The relatively small performance gap from 512 to 1024 suggests that HiDe can function effectively under practical batch size constraints.

These results highlight HiDe’s core advantage: it leverages sample-level variation to extract shared semantics in long-form text, contributing to improved cross-modal alignment even without architectural changes to the encoder.

\begin{table}[t]
\centering
\small
\setlength{\tabcolsep}{2.0pt}
\caption{\textbf{Effect of batch size on HiDe-driven performance.} Larger batches provide more semantic variation for HiDe, leading to improvements in both retrieval (Urban1k, Docci, Long-DCI) and monotonicity metrics (HiMo@2/3/K). H2/3.: HiMo@2/3; HK.: HiMo@K on HiMo-Docci.}
\label{tab:batch_size_comparison}
\begin{tabular}{@{}c *{6}{r} r r r@{}}
\toprule
\multirow{2}{*}{\textbf{Batch Size}} 
& \multicolumn{2}{c}{\textbf{Urban1k}} 
& \multicolumn{2}{c}{\textbf{Docci}} 
& \multicolumn{2}{c}{\textbf{Long-DCI}} 
& \multirow{2}{*}{\textbf{H2.}} 
& \multirow{2}{*}{\textbf{H3.}} 
& \multirow{2}{*}{\textbf{HK.}} \\
\cmidrule(lr){2-3} \cmidrule(lr){4-5} \cmidrule(lr){6-7}
& \textbf{I2T} & \textbf{T2I} 
& \textbf{I2T} & \textbf{T2I} 
& \textbf{I2T} & \textbf{T2I} & & & \\
\midrule
256  & 92.5 & 91.9 & 79.9 & 81.8 & 58.8 & 58.4 & 97.3 & 62.7 & 0.87 \\
512  & 92.9 & 92.1 & 80.9 & 83.3 & 60.6 & 60.1 & 97.5 & 63.1 & 0.87 \\
1024 & \textbf{93.0} & \textbf{93.1} & \textbf{82.4} & \textbf{84.4} & \textbf{62.2} & \textbf{61.9} & \textbf{97.9} & \textbf{64.2} & \textbf{0.88} \\
\bottomrule
\end{tabular}
\end{table}

\begin{table}[t]
\centering
\setlength{\tabcolsep}{6pt}
\renewcommand{\arraystretch}{0.9}
\caption{\textbf{Semantic Stability Index (SSI) of various methods.} Lower is better. * denotes reimplementation.}
\vspace{-0.3cm}
\resizebox{\columnwidth}{!}{
\begin{tabular}{l c c c c c}
\toprule
\textbf{Metric} & Long-CLIP & TULIP & FineLIP\textsuperscript{*} & FG-CLIP & \textbf{HiMo-CLIP (Ours)} \\
\midrule
SSI $\downarrow$ & 11.45 & 12.99 & 8.72 & 10.89 & \textbf{4.63} \\
\bottomrule
\end{tabular}
}
\vspace{-0.4cm}
\label{tab:robustness}
\end{table}

\begin{figure*}[t]
    \centering
    \includegraphics[width=0.9\linewidth]{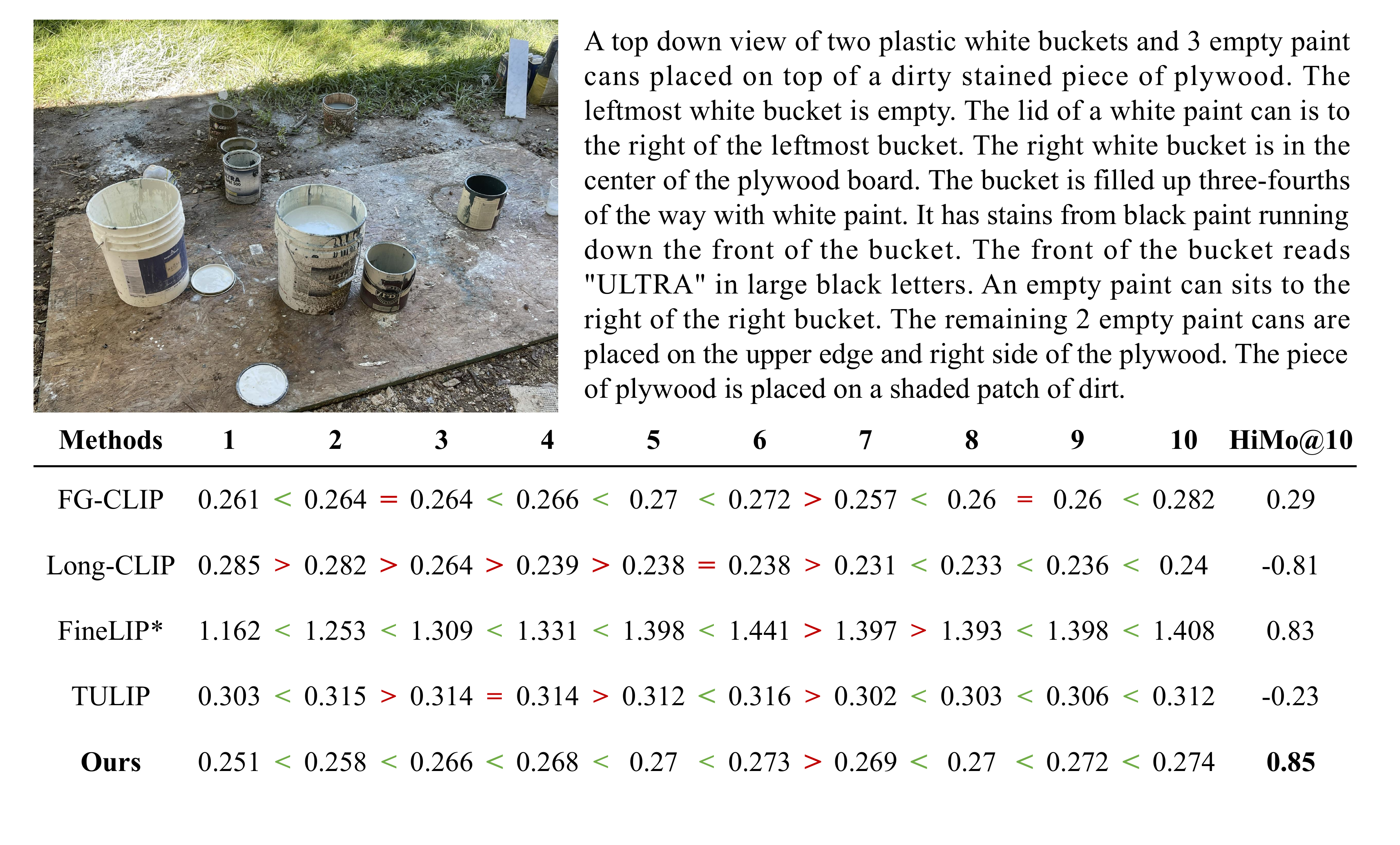}
    \caption{
    \textbf{Bad Case Analysis of Semantic Monotonicity ($K=10$).}  
    Despite occasional violations at specific steps, HiMo-CLIP demonstrates a stable overall increase in similarity as semantic information accumulates. Baselines suffer from score drops due to redundant or ungrounded textual details, such as the label “ULTRA,” which appears in a visually noisy region. HiMo-CLIP mitigates this by filtering out weak semantic cues while preserving alignment consistency.
    }
    \label{fig:badcase}
    \vspace{-0.3cm}
\end{figure*}

\subsection*{C.3 Robustness Evaluation of HiMo-CLIP}
\label{sec:robustness}

\noindent\textbf{Motivation.}  
Robustness to irrelevant or noisy text is essential for vision-language models in real scenarios. We assess HiMo-CLIP’s stability by injecting unrelated sentences—manually verified as semantically irrelevant—into captions from the HiMo-Docci dataset at random positions.

\medskip
\noindent\textbf{Evaluation Metric.}  
We propose the \textit{Semantic Stability Index (SSI)} to measure relative changes in image-text similarity before and after noise insertion:

\begin{equation}
\text{SSI} = \frac{1}{N} \sum_{i=1}^N \frac{1}{M_i} \sum_{j=1}^{M_i} \left| \frac{s^{(i)}_{j,\text{ori}} - s^{(i)}_{j,\text{noise}}}{s^{(i)}_{j,\text{ori}}} \right| \times 100 .
\label{eq:semantic_stability_index}
\end{equation}

where $N$ is the number of samples, $M_i$ the number of hierarchical subtexts, and $s^{(i)}_{j,\text{ori}}$, $s^{(i)}_{j,\text{noise}}$ the similarity scores before and after noise. Lower SSI indicates stronger robustness.

\medskip
\noindent\textbf{Results.}  
Table~\ref{tab:robustness} shows HiMo-CLIP achieves the lowest SSI (\textbf{4.63}), significantly outperforming baselines. This demonstrates its superior resilience to semantic noise, thanks to HiDe’s adaptive semantic decomposition and MoLo’s monotonicity-aware alignment objectives. More examples are shown in Figure~1.


\subsection*{C.4 Bad Case Study: Violation and Robustness of Semantic Monotonicity}

Although HiMo-CLIP encourages semantic monotonicity through its hierarchical representation design, strict monotonic increase in similarity scores cannot always be guaranteed. As shown in Figure~\ref{fig:badcase}, we construct a step-by-step text enrichment case with $K=10$ semantic levels and observe how similarity scores evolve across different models.

Most baseline models (e.g., FG-CLIP, Long-CLIP, FineLIP*, TULIP) exhibit non-monotonic behavior—some even show clear score drops when additional details are added. A prominent decline occurs at the 7\textsuperscript{th} sentence involving the bucket label “ULTRA,” which is visually ambiguous due to occlusion. In contrast, HiMo-CLIP maintains a consistent upward trend, demonstrating its ability to suppress ungrounded or noisy semantics while preserving core alignment.

\subsection*{C.5 HiMo-Docci Dataset Construction}
\label{sec:HiMoDocci dataset}

To evaluate alignment under deep semantic hierarchies, we construct the \textbf{HiMo-Docci} dataset by selecting 1,000 long-form image-text pairs from the Docci dataset. Each caption is manually segmented into semantically progressive subtexts, capturing a gradual increase in descriptive detail.
All segmentations are reviewed by expert annotators to ensure semantic validity, logical coherence, and consistency with the visual content. The final dataset supports reliable benchmarking of hierarchical alignment and monotonicity in long-form retrieval.
Due to file size limits in the supplementary material, we include only the segmented text annotations (see \texttt{himo\_docci\_data.json} in the attachment). The corresponding image files will be made available via an open-access repository upon publication.

\end{document}